\documentclass[letterpaper, 10 pt, conference]{ieeeconf}  

\IEEEoverridecommandlockouts                              

\usepackage{graphics} 
\usepackage{epsfig} 
\usepackage{mathptmx} 
\usepackage{times} 
\usepackage{amsmath} 
\usepackage{amssymb}  
\usepackage{graphicx}
\usepackage{makecell}
\usepackage{caption}
\usepackage{mathrsfs}
\usepackage[ruled,vlined]{algorithm2e}
\usepackage[font={small}]{caption}
\usepackage[mathscr]{euscript}
\usepackage{bm}
\usepackage{pifont}
\usepackage{siunitx}

\usepackage{array}
\newcommand{\PreserveBackslash}[1]{\let\temp=\\#1\let\\=\temp}

\newcolumntype{C}[1]{>{\PreserveBackslash\centering}p{#1}}
\newcolumntype{R}[1]{>{\PreserveBackslash\raggedleft}p{#1}}
\newcolumntype{L}[1]{>{\PreserveBackslash\raggedright}p{#1}}

\usepackage[hidelinks]{hyperref}
\hypersetup{
    colorlinks=true,
    linkcolor=blue,
    urlcolor=blue,
    citecolor=blue
}

\usepackage{balance}

\usepackage[utf8]{inputenc}
\usepackage[english]{babel}
\usepackage[backend=biber,sorting=none,citestyle=numeric-comp]{biblatex}

\title{\LARGE \bf

Generating Humanoid Multi-Contact through Feasibility Visualization

}

\author{Stephen McCrory$^{1,2}$, Sylvain Bertrand$^{1}$, Achintya Mohan$^{3}$, Duncan Calvert$^{1,2}$, Jerry Pratt$^{4}$, Robert Griffin$^{1,2}$ 
\thanks{This work was supported through ONR Grant No. N00014-19-1-2023 and NASA Grant No. 80NSSC20M0197.}%
\thanks{$^{1}$Author is with the Institute of Human and Machine Cognition (IHMC),
        40 S Alcaniz St, Pensacola, FL 32502, USA}
\thanks{$^{2}$Author is with the University of West Florida (UWF),
        11000 University Pkwy, Pensacola, FL 32514, USA}%
\thanks{$^{3}$ Author is with Georgia Institute of Technology, North Avenue Atlanta, GA 30332, USA}%
\thanks{$^{4}$ Author is with Figure AI, Inc., Sunnyvale, CA}%
\thanks{Email: {\tt\small \{smccrory, sbertrand, dcalvert, jpratt, rgriffin\}@ihmc.org, achintya@gatech.edu}}
}

\begin{document}
\maketitle
\thispagestyle{empty}
\pagestyle{empty}

\begin{abstract}

We present a feasibility-driven teleoperation framework designed to generate humanoid multi-contact maneuvers for use in unstructured environments. Our framework is designed for motions with arbitrary contact modes and postures. The operator configures a pre-execution preview robot through contact points and kinematic tasks. A fast estimation of the preview robot’s quasi-static feasibility is performed by checking contact stability and collisions along an interpolated trajectory. A visualization of Center of Mass (CoM) stability margin, based on friction and actuation constraints, is displayed and can be previewed if the operator chooses to add or remove contacts. Contact points can be placed anywhere on a mesh approximation of the robot surface, enabling motions with knee or forearm contacts. We demonstrate our approach in simulation and hardware on a NASA Valkyrie humanoid, focusing on multi-contact trajectories which are challenging to generate autonomously or through alternative teleoperation approaches.

\end{abstract}

\section{INTRODUCTION}

Humanoid robots are frequently designed to be capable of a wide range of motions. Making full use of these platforms as well as understanding their limitations requires an ability to generate and deploy coordinated, multi-contact maneuvers. Generating such motions is achieved by relaxing assumptions about the robot's posture, contactable limbs and contact modes (e.g. planar, line or point contacts) in order to maintain motion generality. The quasi-static case has been well-studied and mature tools exist for reasoning about arbitrary friction constraints \cite{bretl2008testing}, admissible trajectory timing \cite{pham2014general}, and actuation margins \cite{orsolino2020feasible}. However, leveraging these tools for search-based planning presents a number of obstacles. Primarily, multi-contact motions have an inherently large branching factor which can cause planning to be computationally intractable or rely on heuristics \cite{tonneau2018efficient, bouyarmane2018multi}. Additionally, it is challenging to construct generic posture scoring functions that perform well over a large space of configurations. In this work we present a teleoperation framework which builds on existing multi-contact feasibility metrics, namely \cite{bretl2008testing, orsolino2020feasible}, by providing feasibility visualization to the operator. The operator guides the trajectory using contact and posture tasks and is informed of motion feasibility during teleoperation.





\begin{figure}[t]
    \centering
    \includegraphics[width=\columnwidth]{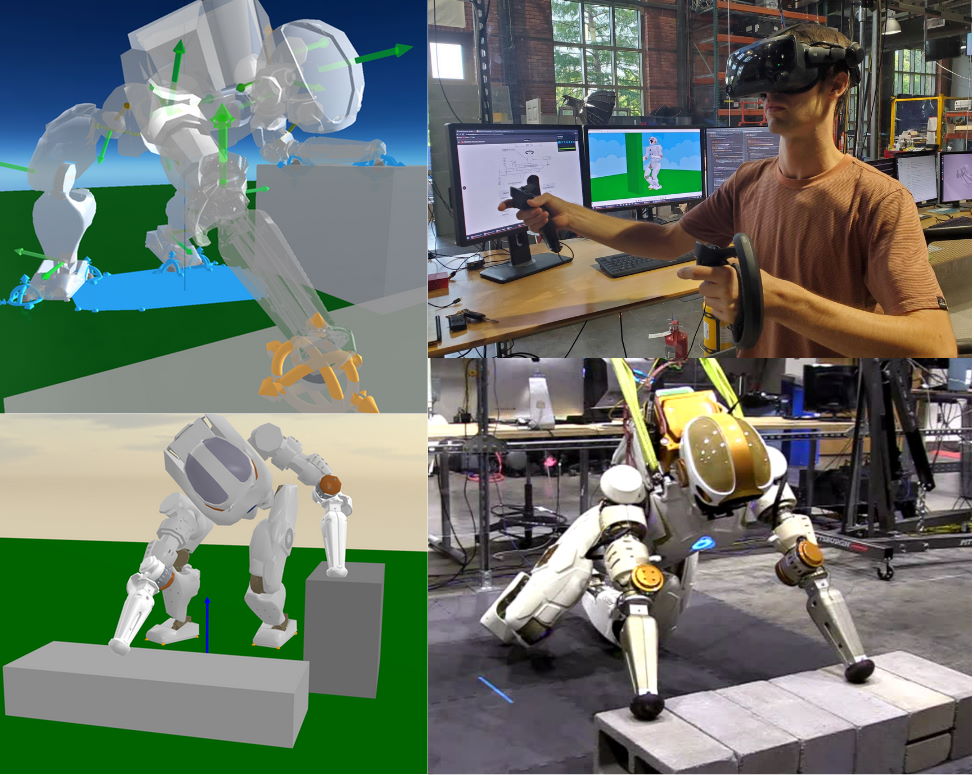}
    \caption{Demonstration of the presented teleoperation framework on Valkyrie in various multi-contact scenarios. The operator (top-right) uses a VR application (top-left) to iteratively create keyframes by dragging a preview robot to the desired configuration. Trajectories are validated in simulation (bottom-left) and hardware (bottom-right) on the Valkyrie humanoid.}
    \label{fig:scripting_title_figure}
\end{figure}

The aim of this teleoperation approach is to capture a wide family of multi-contact maneuvers, including crawling, kneeling, bracing against a wall as well as normal standing. In addition to the contact and actuation feasibility checks mentioned above, these motions require a flexible interface for generating atypical contacts and postures. In our interface, the operator can generate contact points anywhere on the surface of a mesh approximation of the robot. Similarly, taskspace posture setpoints can be generated for any link, with configurable priority weighting and constrained axes. 

The presented work is implemented as a Virtual Reality (VR) interface.  Our design prioritizes flexible operator input and visualizing spatial artifacts, matching many strengths of VR. Research has shown VR offers flexible and expressive operator input when manipulating virtual artifacts \cite{groechel2022tool} and has been shown to reduce operation time compared to conventional desktop interfaces \cite{sheetz2022comparing}. However, our approach is not unique to VR and the presented results are independent of the use of VR. We validate our interface in three simulated teleoperation experiments in which the robot performs multi-contact maneuvers. For experiment two, we deploy the motion on a physical Valkyrie robot to perform a crouch-to-kneel motion.

\renewcommand{\arraystretch}{1.2}
\begin{table*}
\centering
\captionof{table}{Comparison of approaches to motion configurability in multi-contact motion generation systems.} \label{tab:comparison} 
\begin{tabular}{ | c | c | c | c | c | } 
 \hline
 Related Work & Contactable Links & Commanded Posture & Contact Points on Robot & Contact Modes \\
 \hline
 \hline
 \textbf{Ours} & Hands, Elbows, Knees, Feet & Taskspace, Joint Position, CoM & Operator Specified & Point, Line, Plane \\
 \hline
 Brossette et al. \cite{brossette2018multicontact} & Hands, Feet & Taskspace, Joint Position, CoM & Predefined & Point, Plane \\
 \hline
 Rouxel et al. \cite{rouxel2022multicontact} & Hands, Feet & Taskspace & Predefined & Point, Plane \\
 \hline
 Hiraoka et al. \cite{hiraoka2021online} & Hands, Knees, Feet & Taskspace & Operator Specified & Point, Plane \\
 \hline
 Henze et al. \cite{henze2017multi} & Hands, Knees, Feet & Taskspace, Joint Position, CoM & Predefined & Point, Plane \\
 \hline
 Polverini et al. \cite{polverini2020multi} & Hands, Feet & Taskspace, CoM & Predefined & Point \\
\hline
 Otani et al. \cite{otani2017adaptive} & Hands, Elbows, Knees, Feet & Taskspace & Predefined & Point, Plane \\
 \hline
\end{tabular}
\end{table*}

\section{RELATED WORK}

Translating an operator's intent to robot motion is a challenging task, particularly when commanding coordinated motions to legged or dexterous robots. This challenge has given rise to various retargeting frameworks for efficiently mapping operator input to robot motion. Methods for retargeting depend on factors such as available human measurements, task application and control scheme \cite{darvish2023teleoperation}.

There has been significant progress in supervisory-style interfaces for humanoid teleoperation in which the operator provides intermittent, high-level input. This became a popular approach in the DARPA Robotics Challenge (DRC) by having the user provide walking, posture or grasp setpoints with varied levels of customization among the teams \cite{johnson2017team}. For example, Zucker et al. \cite{zucker2015general} included one user-adjustable setpoint per end-effector with jointspace, world-frame and body-frame control modes. Marion et al. \cite{marion2018director} had additional flexibility such as selecting chest, pelvis and CoM setpoints as well as the taskspace constraint set and base of the kinematic chain. Motion validation consisted of collision checks \cite{zucker2015general} and validating that the CoM remains within the support region \cite{marion2018director}.

Adapting these interfaces for multi-contact scenarios presents many challenges, the main one being motion feasibility. For quasi-static motion, a feasible CoM region with respect to contact friction constraints \cite{bretl2008testing} and actuation constraints \cite{orsolino2020feasible} can be employed. This has been extensively used in multi-contact motion planning, particularly for contact-first approaches \cite{polverini2020multi, escande2013planning, vaillant2016multi}. Vaillant et al. \cite{vaillant2016multi} enabled an operator to teleoperate a ladder-climbing scenario by specifying a contact sequence. In contrast to such discrete search-based approaches Brossette et al. \cite{brossette2018multicontact} optimized contact points over a manifold approximation of the contactable environment surface. Rouxel et al. \cite{rouxel2022multicontact} retarget an operator-commanded set of posture and contact constraints to achieve quasi-statically stable multi-contact motions.

Another major challenge with teleoperation in multi-contact scenarios in enabling a high degree of operator expression for commanding arbitrary postures and contact modes. The use of VR and Motion Capture (MoCap) enables more sophisticated operator mapping through partial or whole-body kinematic retargeting. Early pioneering work by Pollard et al. \cite{pollard2002adapting} used MoCap data from a human actor to mimic dancing motions by mapping to the torso and upper body of a fixed-base humanoid. The operator-to-robot mapping accounted for joint limits, velocity limits and singularities. Choi et al. \cite{choi2019towards} used MoCap for retargeting human motion data to humanoid robots with vastly different morphologies. There has also been work in using MoCap for more contact-rich scenarios \cite{di2016multi, otani2017adaptive}. Otani et al. \cite{otani2017adaptive} implemented a motion capture based retargeting scheme for multi-contact manipulation scenarios. Operator motion is mapped to support, free and manipulation tracking sets and demonstrated in simulation by bracing against a table while manipulating an object. 



\begin{figure}[!t]
    \centering
    \includegraphics[width=0.83\columnwidth]{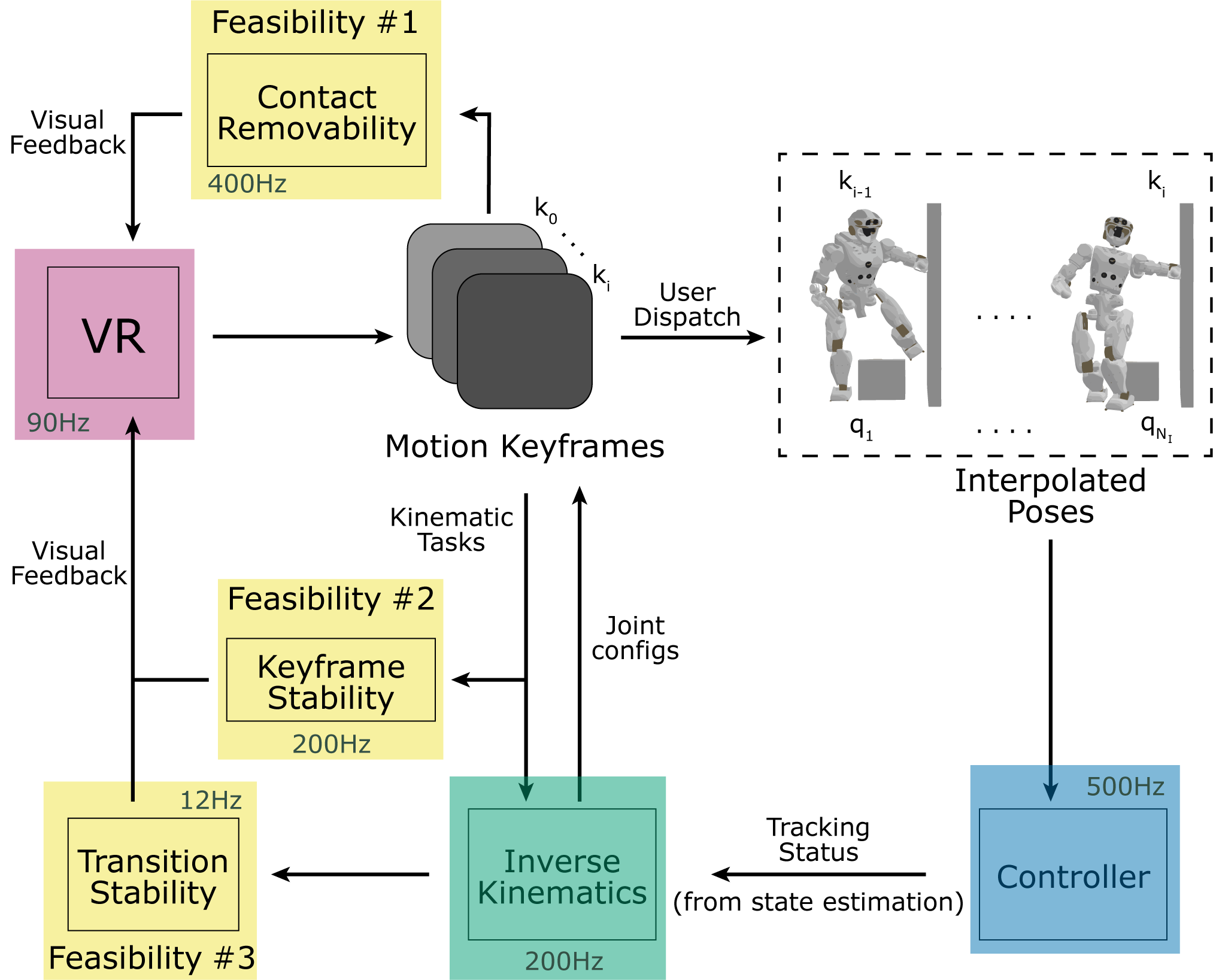}
    \caption{Control flow of our multi-contact teleoperation framework.}
    \label{fig:control_flow}
\end{figure}

\subsection{Contribution}
Our teleoperation framework is designed for high motion configurability by relaxing restrictions on the set of contactable limbs and commanded posture objectives (Tab. \ref{tab:comparison}). The main contribution of this framework are as follows:

\begin{itemize}
\item Teleoperation with contact points placed arbitrarily on the robot surface and contact feasibility visualized as an actuation-aware CoM stability margin.
\item Transition feasibility by visualizing contact removability and checking contact stability along an interpolated trajectory.
\item Interface controls to configure the weight, axis constraints and control frame of posture tasks.
\end{itemize}

\begin{figure*}
    \centering
    \includegraphics[width=\textwidth]{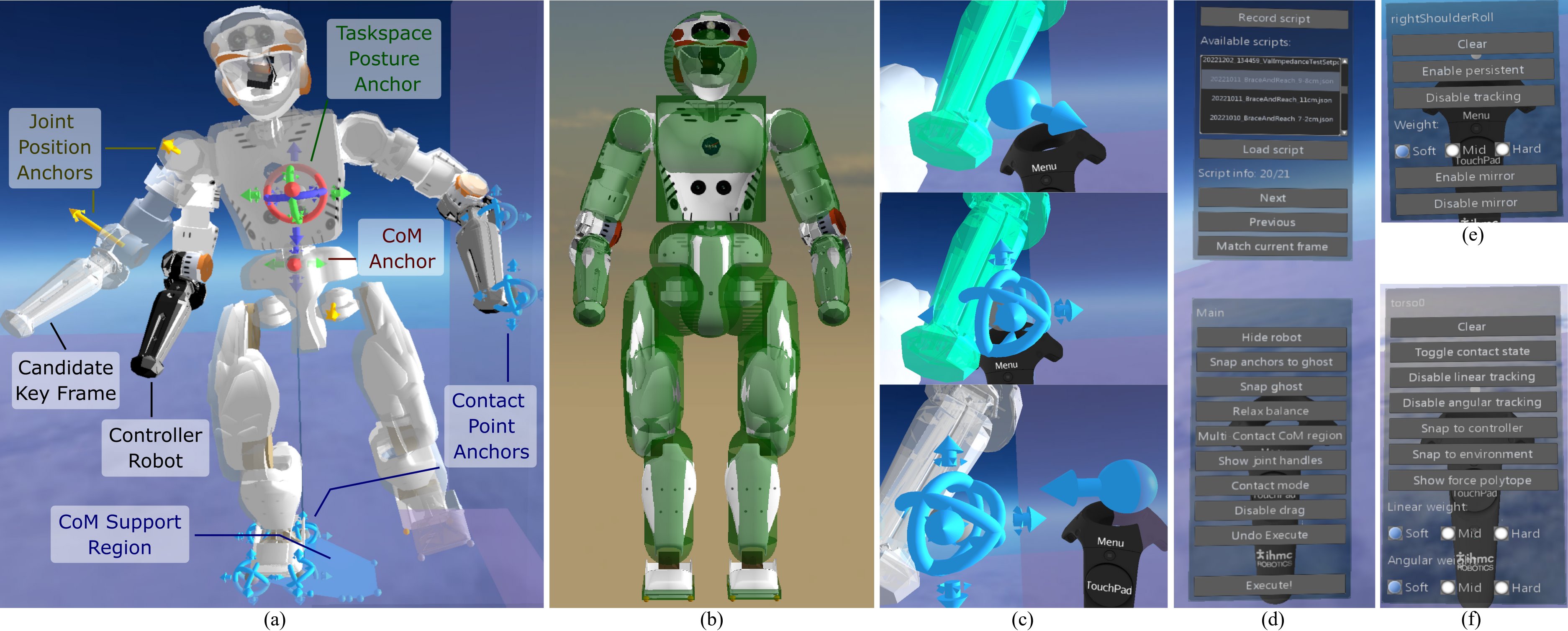}
    \caption{(a) Operator view in VR with Valkyrie leaning against a wall. (b) Convex mesh model of Valkyrie. Taskspace posture anchors are generated with a control frame that lies inside the shape(s) corresponding to the configured link. Similarly, contact point anchors are generated along the surface of the shape(s) corresponding to the contacting link. (c) Contact points are projected to the surface of the robot or environment mesh while creating (top) or placing (bottom) a contact point anchor. The arrow indicates the surface normal of the robot or environment mesh. (d) Main menu used to export/load keyframe scripts, toggle visualization, configure solver behavior and switch between placing taskspace posture and contact anchors. (e) Joint anchor menu. (f) Taskspace posture/contact point anchor menu.}
    \label{fig:teleop_combo}
\end{figure*}



\section{TELEOPERATION FRAMEWORK}

An outline of our framework is shown in Fig. \ref{fig:control_flow}. A \textit{keyframe} represents a whole-body configuration of the robot. At a high level, teleoperation is performed by iteratively generating a keyframe and dispatching it to the controller. Keyframes are generated in VR by configuring a set of kinematic tasks, such as desired contact state, taskspace posture, preferred joint angles and CoM position. An inverse kinematics (IK) solver continuously processes these inputs and computes the candidate keyframe along with an interpolated trajectory. The operator receives real-time feedback based on 3 feasibility metrics: (1) the set of contacts that are currently removable, (2) the static stability and collision status of the keyframe and (3) the static stability and collision status of the interpolated trajectory to the keyframe. The feasibility estimation is then displayed to the user along with relevant data such as CoM stability margin and joint actuation saturation. When a user is finished configuring the candidate keyframe and the feasibility estimation is valid, it can be dispatched to the robot. 


\subsection{Kinematic Task Generation} \label{sec:task_generation}

The VR interface is designed to enable an operator to pose the robot by creating, removing or modifying constraints on-the-fly. To do this, four types of virtual interactable ``anchors'' are available, each corresponding to a type of kinematic task (Fig \ref{fig:teleop_combo}(a)):

\begin{itemize}

    \item \textbf{Taskspace Posture}: Matches a reference frame $F_p$ that is rigidly attached to link $l$ with a desired frame $F_d$ expressed in world coordinates.
    \item \textbf{Center of Mass}: Matches the robot's CoM with a desired position expressed in world coordinates.
    \item \textbf{Joint Position}: Matches to a preferred joint angle.
    \item \textbf{Contact Point}: Matches a contact point $p_c$ that is rigidly attached to link $l$ with a desired position $p_d$ in world coordinates.
 
\end{itemize}


An approximate convex decomposition of the robot's mesh (Fig. \ref{fig:teleop_combo}(b)) is used to aid the link-VR controller association. When the VR controller enters a convex shape of the preview robot, the corresponding link $l$ is highlighted and the user can click to generate a taskspace control frame $F_p$ at the VR controller's pose, which is initially coincident with $F_d$. The virtual anchor, which represents $F_d$, can then be dragged and rotated to the desired pose. The user can toggle which of the three linear and angular constraint axes are enabled, reflected visually with highlights on the anchor (Fig. \ref{fig:teleop_combo}(a)).

The CoM anchor is created by selecting a marker designating the candidate keyframe's current CoM. Similar to the taskspace posture anchor, the user can enable which linear axes are constrained. Joint position anchors can be enabled for any joint and are primarily used to bias the IK solver within a nullspace. All kinematic tasks are assigned a relative priority level by specifying one of soft, mid or hard weights, which are set through anchor menus (Fig. \ref{fig:teleop_combo}(e,f)). The option to ``Snap anchors to ghost'' (Fig. \ref{fig:teleop_combo}(d)) will move the setpoint of all non-contact kinematic tasks to the currently achieved IK configuration.

\subsection{Contact Points}
Contact point anchors are created in a specific mode which projects the VR controller positions to the surface of the nearest convex shape of the preview robot (Fig. \ref{fig:teleop_combo}(b)), with an arrow showing the surface normal. Environment shapes are also modelled as convex shapes. Contact points can be snapped to the environment using a similar mode that projects the VR controller to the surface of the nearest environment shape and shows the environment surface normal. Fig. \ref{fig:teleop_combo}(c)-top shows a contact point being created on the robot's forearm and Fig. \ref{fig:teleop_combo}(c)-bottom shows the contact point being snapped to a wall. Line and plane contacts are created by configuring the set of contact points accordingly.

\section{Kinematics Solver} \label{sec:ik}

An optimization-based IK solver is used to compute quasi-statically stable whole-body configurations given a set of kinematic tasks. We solve the IK problem using Sequential Quadratic Programming (SQP) due to its generality and success on humanoids \cite{brossette2018multicontact, rouxel2022multicontact, beeson2015trac}. At every solve step, a desired velocity $\mathbf{v}_d\in\mathbb{R}^{n+6}$ is computed to drive the model towards a configuration that achieves the desired task objectives, where $n$ is the number of actuated degrees of freedom in the robot. Note that $\mathbf{v}_d$ represents the velocity of the solver model and is independent of the controller's velocity. The IK iteratively solves the following Quadratic Program (QP): \begin{equation}
\begin{aligned}
 \label{eq:IKQP}
    \min_{\mathbf{v}_d} \quad   & c_{\mathrm{nom}} + c_{\mathbf{J}} + c_{\mathbf{v}_d}    \\
    \textrm{s.t.} \quad         & \mathbf{v}_{min} \leq \mathbf{v}_d \leq \mathbf{v}_{max}      
\end{aligned}
\end{equation}

The objective function terms are given by: 

\renewcommand{\arraystretch}{1.35}
\setlength{\tabcolsep}{2.5pt}

\begin{tabular}{ll}
 Nominal Objective: & $         c_{\mathrm{nom}}    = (\mathbf{v}_d - \mathbf{v}_{\mathrm{nom}})^T  \mathbf{C}_{\mathrm{nom}}    (\mathbf{v}_d - \mathbf{v}_{\mathrm{nom}}) $ \\
 Kinematic Tasks: & $         c_{\mathbf{J}}      = (\mathbf{J}\mathbf{v}_d - \mathbf{p})^T       \mathbf{C}_{\mathbf{J}}      (\mathbf{J}\mathbf{v}_d - \mathbf{p}) $ \\
 Velocity Cost: & $   c_{\mathbf{v}_d}      = \mathbf{v}_d^T \mathbf{C}_{\mathbf{v}_d} \mathbf{v}_d$,
\end{tabular}

where the terms are given by:

\begin{itemize}
    \item $\mathbf{v}_{\mathrm{nom}}$ drives the robot to a nominal whole-body configuration, which by default is the controller's current configuration. The user can set the nominal configuration as IK's current solution as by selecting ``snap ghost'' (Fig. \ref{fig:teleop_combo}(d)), which is generally used for larger motions where the controller and IK differ significantly.
    \item $\mathbf{J} = [\mathbf{J}^T_1 \ldots \mathbf{J}^T_k]^T$ and $\mathbf{p} = [\mathbf{p}^T_1 \ldots \mathbf{p}^T_k]^T$ are the stacked Jacobian matrices and motion objectives computed as feedback terms from the kinematic tasks and $\mathbf{C_J} = \mathrm{diag}(\mathbf{w}_0, \ldots, \mathbf{w}_k)$ is a block-diagonal weight matrix, which is detailed below.
    \item $\mathbf{v}_{min}$ and $\mathbf{v}_{max}$ bound the joint velocity so the joint remains within its bounds within the update period $\Delta T$.
    \item $\mathbf{C}_{\mathrm{nom}} = 0.5\,\mathbf{I}_{n+6}$ and $\mathbf{C}_{\mathbf{v}_d} = 0.1\,\mathbf{I}_{n+6}$ are constant weight matrices.
\end{itemize}

With each solve iteration, the candidate keyframe configuration $\mathbf{q}_d$ is updated by integrating the computed velocities (Eq. \ref{ikupdate}), where $\Delta T$ is the solver update period. \begin{equation} \label{ikupdate}
    \mathbf{q}_d \leftarrow \mathbf{q}_d + \mathbf{v}_d \Delta T
\end{equation}

For kinematic task $i$, a task Jacobian $\mathbf{J}_i$ and feedback motion $\mathbf{p}_i$ are used to compute a feedback term:

\begin{itemize}
    \item Joint Position: $\mathbf{J}_i$ is a selection matrix for the joint and $\mathbf{p}_i$ is a velocity proportional to the position error.
    \item Center of Mass: $\mathbf{J}_i$ is the linear centroidal momentum matrix $\mathbf{A}$ \cite{orin2008centroidal} and the objective $\mathbf{p}_i$ is a momentum proportional to the linear position error.
    \item Taskspace Posture and Contact Point: $\mathbf{J}_i$ is the geometric Jacobian \cite{spong2020robot} of the control frame $F_p$ (Sec. \ref{sec:task_generation}). A proportional feedback law on the relative transform between $F_p$ and $F_d$ is used to compute $\mathbf{p}_i$ \cite{bullo1995proportional}.
\end{itemize}

Kinematic tasks are each assigned a weight matrix $\mathbf{w}_i = w_i \mathbf{I}_{N_i}$, where $w_i$ is computed from the task's priority given by Tab. \ref{tab:weights} and $N_i$ is the tasks's dimensionality. Note that the CoM task weight is scaled down by the robot mass $m$ so feedback is only dependent on kinematic quantities. For Taskspace Posture, Center of Mass and Contact Points, a selection matrix $\mathbf{S}_i$ is used to only provide feedback for the constrained axes. This is done by premultiplying both the Jacobian and objective by $\mathbf{S}_i$.

\renewcommand{\arraystretch}{1.2}
\begin{table}[t!]
\centering
\captionof{table}{Kinematic Task Weights} \label{tab:weights} 
\begin{tabular}{ c c c c } 
 \hline
 & Soft & Mid & Hard \\ 
 \hline
Taskspace Posture & 0.1 & 1.0 & 10.0 \\ 
CoM & $0.01/m$ & $0.1/m$ & $1.0/m$ \\ 
Joint Position & 0.1 & 1.0 & 10.0 \\ 
Contact Point & 50.0 & 200.0 & 500.0 \\ 
 \hline
\end{tabular}
\end{table}

\begin{figure*}
    \centering
    \includegraphics[width=\textwidth]{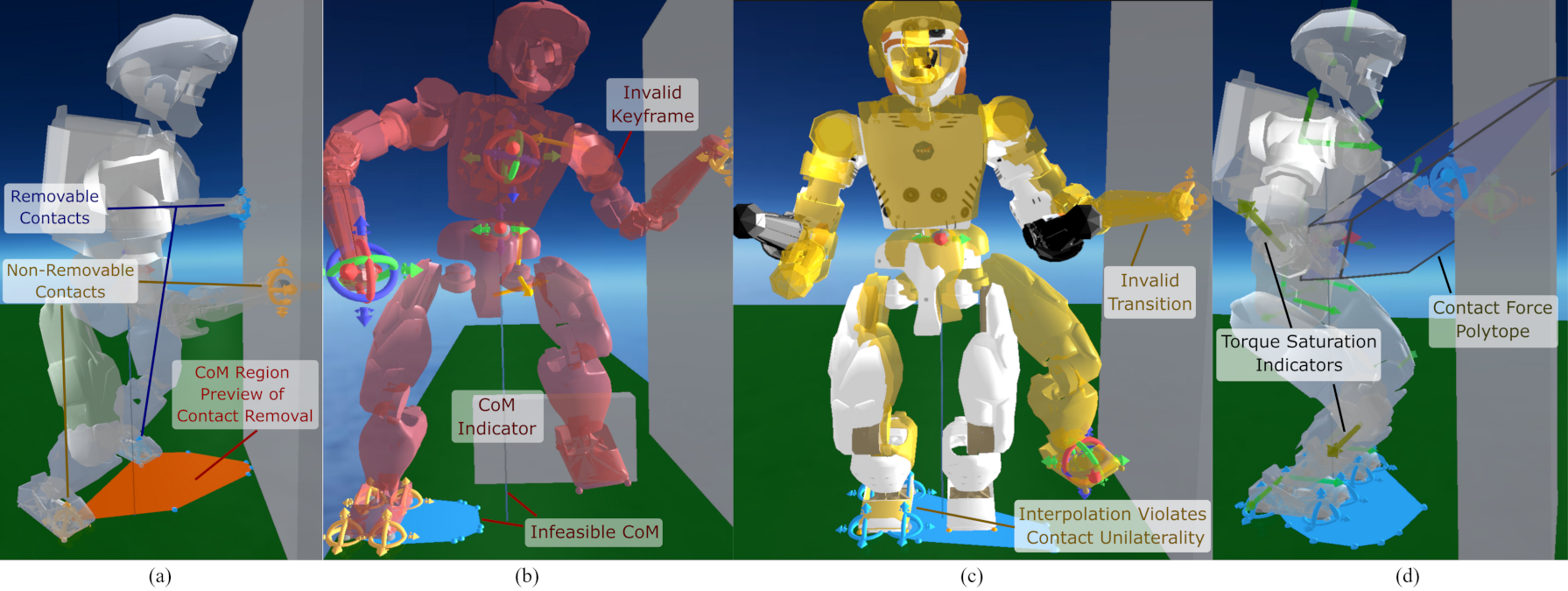}
    \caption{(a) Contact anchor color indicates whether a contact is removable. The operator can select a contact and preview the CoM region in the absence of the contact. The CoM region shown is a preview with the right arm contact removed, which is currently infeasible. The preview robot turns red if a keyframe is invalid (b) or yellow if the keyframe transition is invalid. In (c), although the keyframe is valid, the keyframe transition requires the CoM to leave the CoM feasible region before placing the left hand. (d) Actuation feasibility can be visualized by a force polytope at contact points and by indicating joint torque saturation. In this figure the right knee and elbow are a darker shade, indicating they are close to torque saturation.}
    \label{fig:feasibility}
\end{figure*}

\section{Feasibility Estimation} \label{sec:feasbility}


During teleoperation, motion feasibility is assessed for the candidate keyframe and keyframe transition. In \ref{sec:com_feas} we present our approach for assessing the CoM stability margin \cite{bretl2008testing, orsolino2020feasible} based on the preview robot's state. In \ref{sec:interpolation} we present our method for anchor-based kinematic interpolation. In \ref{sec:feasibility_visualization} we then present our approach for visual feasibility cues which are based on three motion feasibility checks: contact removal feasibility, keyframe configuration feasibility and keyframe transition feasibility.

\subsection{CoM Stability Margin} \label{sec:com_feas}

The set of contact point anchors describes the candidate keyframe's contact state. A contact anchor $i$ is parameterized by a contact point position $\mathbf{r}_i$, contact normal $\mathbf{n}_i$ and contact force $\mathbf{f}_i$. The jointspace rigid body equations of motion \cite{featherstone2014rigid} for the quasi-static case of $\mathbf{\ddot{q}} \approx \mathbf{\dot{q}} \approx \mathbf{0}$ are given by:
\begin{equation}\label{eq:static_eq_eom}
    \mathbf{G} - \boldsymbol{\tau} = \sum_{i=1}^{N_c} \mathbf{J}^T_{c,i} \mathbf{f}_i = \mathbf{J}^T_c \mathbf{f} \,.
\end{equation} 
Where $N_c$ is the number of contact points, $\mathbf{G}$ is the gravitational torque vector, $\boldsymbol{\tau}$ are the set of joint torques, and $\mathbf{J}_{c,i}\in \mathbb{R}^{3\times n}$ is the Jacobian for contact $i$. A stacked contact Jacobian $\mathbf{J}_c$ and force vector $\mathbf{f}$ are used for brevity. For a given keyframe, static equilibrium is expressed as the feasibility problem:
%

%
\begin{equation} \label{eq:StaticFeasibility}
\begin{aligned}
    \exists \, \mathbf{f} \;\; \textrm{s.t.}   \quad &  \mathbf{f}_i \in \mathscr{K}_i  \\
                                            \quad &   \sum_i \mathbf{f}_i = -m\mathbf{g} \\ 
                                            \quad &   \sum_i \mathbf{r}_i \times \mathbf{f}_i = -\mathbf{c} \times (m\mathbf{g}) \\
                                            \quad &   \mathbf{G} - \boldsymbol{\tau}^{+} \leq \mathbf{J}^T_c\mathbf{f} \leq \mathbf{G} - \boldsymbol{\tau}^{-},
\end{aligned}
\end{equation}
where $\mathscr{K}_i$ is the friction cone of contact $i$, $\mathbf{g} = (0,0,-9.81)^T$ is gravitational acceleration, $\mathbf{c}$ is the CoM position and $\boldsymbol{\tau}^{-}, \boldsymbol{\tau}^{+}$ are the lower and upper joint torque bounds. The standard linearized friction model \cite{bouyarmane2018multi} is used so that Eq. \ref{eq:StaticFeasibility} contains only linear constraints. Using this set of linear constraints, we compute the preview robot's CoM stability margin using the Iterative Projection algorithm introduced by Bretl et al. \cite{bretl2008testing}. This approach recursively solves a Linear Program to compute maximal CoM displacements along a set of query directions (see Appendix for details on region calculation). If the preview robot's CoM is inside this region, it indicates the posture is quasi-statically stable with respect to friction and actuation constraints.




\subsection{Kinematic Interpolation} \label{sec:interpolation}


Kinematic interpolation is done by computing a set of intermediate whole-body configurations that smoothly blend the kinematic task sets between two consecutive keyframes $K_0$ (start) and $K_1$ (end). To do this, each kinematic task is assigned a corresponding task on the opposite side of interpolation. While this is trivial for tasks expressed in both $K_0$ and $K_1$, we create additional ``placeholder'' tasks for those only present in $K_0$ or $K_1$. The placeholder task corresponding to task $T$ is given zero weight and a setpoint equal to $T$ expressed in the opposite keyframe. For example, if a taskspace posture task for the hand is present in $K_0$ but not $K_1$, the placeholder setpoint is the frame $F_p$ on the hand expressed in $K_1$.

The intermediate configurations are computed by interpolating along discrete, evenly-spaced points parameterized by a phase variable $s = \frac{i}{N_I+1}, i = \{1, 2, ..., N_I\}$. Kinematic task weights are interpolated using the formula:
\begin{equation}
    w(s) = w_0 (1-s)^{\alpha_0} + w_1 s ^ {\alpha_1},
\end{equation}
where $w_0$ and $w_1$ are the kinematic task's weight at the start and end of interpolation and $\alpha_0$, $\alpha_1$ are tuning parameters. Kinematic task setpoints are interpolated linearly for spatial and joint positions. Orientation setpoints are interpolated using Spherical Linear Interpolation (Slerp) \cite{shoemake1985animating}. Additionally, for joints that do not contain a nominal task we create a joint position task to bias the start and end of interpolation to match the keyframe configurations with a fixed bias weight of $w_b=1.5$ and tuning exponent $\alpha=6$. In all other cases, weights are interpolated linearly with $\alpha=1$.

We find that a value $N_I=12$ is sufficient for validating transitions between consecutive keyframes due to sufficient resolution while still yielding a fast computation time of 77ms (see Tab. \ref{tab:timing}). For $N_I=12$, we observe an average deviation of 2mm for contacting links and 4mm for non-contacting links when compared to interpolating at significantly higher values of $N_I$. 

Contact switches occur at either the start or end of interpolation. Therefore, the active set of contact points during interpolation are contact points present in both $K_0$ and $K_1$.

\begin{figure*}
    \centering
    \includegraphics[width=\textwidth]{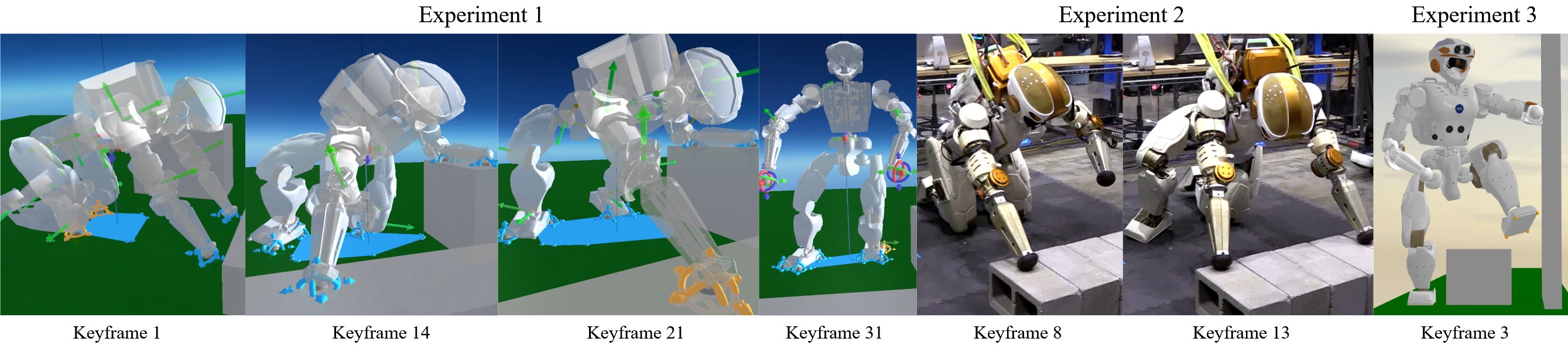}
    \caption{Experiment 1 (simulation): the robot was teleoperated from an initial crouching stance to standing with handholds available in front and to the left of the robot. Experiment 2 (hardware): a modification of the experiment 1 script was deployed to a physical Valkyrie robot. The robot places both hands on the cinder blocks and lifts the right knee to place the right foot on the ground. Experiment 3 (simulation): the robot braces against a wall using the forearm and swings the left foot over an obstacle.}
    \label{fig:results_combo}
\end{figure*}

\subsection{Feasibility Visualization} \label{sec:feasibility_visualization}

To compute if a contact is removable, a modified feasible CoM region is computed with the given contact removed. A contact is considered removable only if the CoM is contained in the modified feasible region. Removability is continuously updated for all contacts as the keyframe configuration changes and non-removable contacts are marked visually and cannot be removed. The operator can select a contact and preview the CoM feasible region if it were removed in order to adjust the CoM accordingly before removing a contact as shown in Fig. \ref{fig:feasibility}(a).

The candidate keyframe and the $N_I$ interpolated configurations are checked for kinematic and static feasibility. Each of the $(1 + N_I)$ configurations has the following checks:

\begin{itemize}
    \item Kinematics-Statics solver (Eq. \ref{eq:IKQP}) cost converges such that successive costs are within a bound $\epsilon_{ik} = 10^{-5}$.
    \item Contact point anchors track with a bound $\epsilon_c=10^{-3}$\si{\meter}.
    \item The feasible CoM region contains the CoM, indicating friction and actuation constraints are statically achieved.
    \item Configuration has no environment collisions among non-contacting links.
\end{itemize}

The preview robot color is modified to reflect whether the candidate keyframe or transition is infeasible. In the event the preview robot is near actuation limits, the operator is also aided by visual cues to determine the joint that is actuation-limited and how to adjust the posture (Fig. \ref{fig:feasibility}(d)). Given the candidate keyframe's contact state, a static force distribution $\mathbf{f}_g$ is computed (see Appendix for details). The corresponding joint torques $\boldsymbol{\tau}_g$ are computed from Eq. \ref{eq:static_eq_eom}:
\begin{equation}\label{eq:static_torques}
    \boldsymbol{\tau}_g = \mathbf{G} - \mathbf{J}^T_c \mathbf{f}_g.
\end{equation}
The color of joint position anchors is updated so that elements of $\boldsymbol{\tau}_g$ near the torque bounds are colored red to alert the operator.
The force polytope $\mathscr{P}_i$ at a contact anchor $i$ can also be visualized, given by \cite{chiacchio1997force}: 
\begin{equation}
    \mathscr{P}_i = \{ \mathbf{f}_i \in \mathbb{R}^3 \; | \; \boldsymbol{\tau}^{-} \leq \mathbf{J}_{c,i}^T \mathbf{f}_i \leq \boldsymbol{\tau}^{+} \} \,.
\end{equation}
This can be useful if the polytope's ``major axis'' has an intuitive preferred orientation, such as vertical when contacting the ground or horizontal when bracing against a wall. 


\section{RESULTS}

\subsection{Framework Timing}

The teleoperation framework divides computation among three threads: a kinematics-statics thread that computes Eq. \ref{eq:IKQP}, a VR thread and a transition feasibility thread. The VR application is run on a Valve Index \cite{Index} at a refresh rate of 90\si{\hertz}. We use Java Monkey Engine \cite{JME} for scene graphics. The execution times of various tasks in the framework were measured and are reported in Tab. \ref{tab:timing}. Each reported time is the average over a window of 100 updates. The evaluation was performed on a desktop with a 10th gen 10-core i9 processor (3.70 \si{\giga\hertz}). This was measured under a nominal operating condition while standing with 8 contact anchors, two taskspace posture anchors, two joint position anchors and a CoM anchor.

\renewcommand{\arraystretch}{1.0}
\begin{table}[h!]
\centering
\captionof{table}{Timing Evaluation} \label{tab:timing} 
\begin{tabular}{ | c | c | c | } 
 \hline
 Task & Description & Time (\si{\milli\second}) \\
 \hline
 \hline
 Kinematics-Statics & \textbf{Total} & \textbf{4.63} \\
        Solver    & CoM Region & 4.05 \\
                           & Kinematics Solver & 0.58 \\
        \hline 
VR Interface & \textbf{Total} & \textbf{5.81} \\
        & VR API & 2.00 \\
        & Kinematic Task Processing & 1.06 \\
        & Collision Check & 0.06 \\
        & Contact Removal & 2.43 \\
        & Gravity Compensation Torques & 0.26 \\
        \hline
Transition Feasibility & \textbf{Total} & \textbf{77.3} \\
        \hline
\end{tabular}
\end{table}

\subsection{Experiment 1: Crouch to Stand, Simulated}

We tested\footnote[2]{A video of the teleoperation is available at \url{https://www.youtube.com/watch?v=AJlSsU4Tvkk}} our teleoperation framework in a simulated scenario where Valkyrie starts in a crouched position on the ground next to two flat boxes, to the front and left of the robot. Tab. \ref{tab:script_stats} shows the teleoperation statistics, including anchor count and achieved contact modes. As shown on the bottom rows, significant use of line contacts was made in addition to point and plane contacts. Fig. \ref{fig:results_combo} highlights important keyframes and contact modes. In keyframe 14, the left forearm is supported through a line contact between the elbow and hand. Also in keyframe 14, the right foot forms a line contact on the inside edge of the foot. In keyframe 21, the front edge of the left foot forms a line contact while placing the foot down. The trajectory was executed using a whole-body impedance controller at 2s per keyframe with an average joint tracking $(0.0043\pm 0.002)$\si{\radian}.

\renewcommand{\arraystretch}{1.0}
\begin{table*}[h!]
\begin{center}
\captionof{table}{Experiment 1 Statistics Simulated Crouch to Stand. Operation Time: 16m18s.} \label{tab:script_stats} 
\begin{tabular}{| c | c | c | c | c | c | c | c | c | c | c | c | c | c | c | c | c | c | c | c | c | c | c | c | c | c | c | c | c | c | c | c | c |}
\hline
 & Keyframe & 1 & 2 & 3 & 4 & 5 & 6 & 7 & 8 & 9 & 10 & 11 & 12 & 13 & 14 & 15 & 16 & 17 & 18 & 19 & 20 & 21 & 22 & 23 & 24 & 25 & 26 & 27 & 28 & 29 & 30 & 31 \\
 \hline
\hline
 & Event & \multicolumn{5}{c|}{\text{Place R. Hand}} & \multicolumn{5}{c|}{\text{Place L. Hand/Elbow}} & \multicolumn{4}{c|}{\text{Place R. Foot}} & \multicolumn{7}{c|}{\text{Lift L. Knee, Place L. Foot}} & \multicolumn{3}{c|}{\text{Lift L. Elbow}} & \multicolumn{3}{c|}{\text{Lift R. Hand}} & \multicolumn{4}{c|}{\text{Lift L. Hand}} \\
 \hline
\rotatebox[origin=c]{90}{Taskspace} & \makecell{Total \\ Modified \\ Add/Rem.} & \makecell{0\\  - \\ - }  & \makecell{1\\  - \\+1}  & \makecell{1\\ 1\\ - }  & \makecell{0\\  - \\-1}  & \makecell{0\\  - \\ - }  & \makecell{1\\  - \\+1}  & \makecell{1\\ 1\\ - }  & \makecell{1\\ 1\\ - }  & \makecell{0\\  - \\-1}  & \makecell{0\\  - \\ - }  & \makecell{1\\  - \\+1}  & \makecell{0\\  - \\-1}  & \makecell{0\\  - \\ - }  & \makecell{0\\  - \\ - }  & \makecell{1\\  - \\+1}  & \makecell{1\\ 1\\ - }  & \makecell{0\\  - \\-1}  & \makecell{0\\  - \\ - }  & \makecell{0\\  - \\ - }  & \makecell{0\\  - \\ - }  & \makecell{0\\  - \\ - }  & \makecell{1\\  - \\+1}  & \makecell{1\\  - \\ - }  & \makecell{0\\-\\-1}  & \makecell{0\\  - \\ - }  & \makecell{0\\  - \\ - }  & \makecell{1\\  - \\+1}  & \makecell{1\\ 1\\ - }  & \makecell{1\\ 1\\ - }  & \makecell{2\\  - \\+1}  & \makecell{2\\ 2\\ - }  \\
 \hline
\rotatebox[origin=c]{90}{Joint} & \makecell{Total \\ Modified \\ Add/Rem.} & \makecell{4\\  - \\ - }  & \makecell{5\\  - \\+1}  & \makecell{5\\ 1\\ - }  & \makecell{5\\  - \\ - }  & \makecell{5\\  - \\ - }  & \makecell{6\\  - \\+1}  & \makecell{5\\  - \\-1}  & \makecell{5\\  - \\ - }  & \makecell{6\\  - \\+1}  & \makecell{6\\ 1\\ - }  & \makecell{6\\  - \\ - }  & \makecell{9\\ 1\\+3}  & \makecell{10\\ 3\\+1}  & \makecell{4\\  - \\-6}  & \makecell{4\\  - \\ - }  & \makecell{4\\  - \\ - }  & \makecell{7\\ 1\\+3}  & \makecell{6\\ 2\\-1}  & \makecell{6\\ 3\\ - }  & \makecell{7\\ 4\\+1}  & \makecell{2\\ 1\\-5}  & \makecell{2\\  - \\ - }  & \makecell{2\\  - \\ - }  & \makecell{3\\-\\+1}  & \makecell{3\\  - \\ - }  & \makecell{3\\  - \\ - }  & \makecell{3\\  - \\ - }  & \makecell{4\\ 1\\+1}  & \makecell{4\\  - \\ - }  & \makecell{4\\  - \\ - }  & \makecell{4\\  - \\ - }  \\
 \hline
\rotatebox[origin=c]{90}{CoM} & \makecell{Enabled \\ Modified } & \makecell{\checkmark\\ \checkmark}  & \makecell{\checkmark\\ \checkmark}  & \makecell{\checkmark\\ \checkmark}  & \makecell{\checkmark\\ \checkmark}  & \makecell{\checkmark\\ \checkmark}  & \makecell{\checkmark\\ \checkmark}  & \makecell{\checkmark\\ \checkmark}  & \makecell{\checkmark\\ \checkmark}  & \makecell{\checkmark\\ \checkmark}  & \makecell{\checkmark\\ \checkmark}  & \makecell{\checkmark\\ \checkmark}  & \makecell{\checkmark\\ \checkmark}  & \makecell{\checkmark\\ \checkmark}  & \makecell{\checkmark\\ \checkmark}  & \makecell{ - \\ \checkmark}  & \makecell{\checkmark\\ \checkmark}  & \makecell{\checkmark\\ \checkmark}  & \makecell{\checkmark\\ \checkmark}  & \makecell{\checkmark\\ \checkmark}  & \makecell{\checkmark\\ \checkmark}  & \makecell{\checkmark\\ \checkmark}  & \makecell{\checkmark\\ \checkmark}  & \makecell{\checkmark\\ \checkmark}  & \makecell{\checkmark\\ \checkmark}  & \makecell{\checkmark\\ \checkmark}  & \makecell{\checkmark\\  - }  & \makecell{\checkmark\\  - }  & \makecell{\checkmark\\ \checkmark}  & \makecell{\checkmark\\ \checkmark}  & \makecell{\checkmark\\ \checkmark}  & \makecell{\checkmark\\ \checkmark}  \\
 \hline
\rotatebox[origin=c]{90}{Contact} & \makecell{Total \\ Modified \\ Add/Rem.} & \makecell{4\\  - \\ - }  & \makecell{3\\  - \\-1}  & \makecell{3\\  - \\ - }  & \makecell{4\\  - \\+1}  & \makecell{4\\  - \\ - }  & \makecell{3\\  - \\-1}  & \makecell{3\\  - \\ - }  & \makecell{3\\  - \\ - }  & \makecell{4\\  - \\+1}  & \makecell{5\\  1 \\+1}  & \makecell{4\\  - \\-1}  & \makecell{4\\  - \\ - }  & \makecell{4\\  - \\ - }  & \makecell{6\\ 2\\+2}  & \makecell{5\\  - \\-1}  & \makecell{6\\  - \\+1}  & \makecell{6\\  - \\ - }  & \makecell{6\\  - \\ - }  & \makecell{6\\  - \\ - }  & \makecell{6\\  - \\ - }  & \makecell{7\\  - \\+2,-1}  & \makecell{6\\  - \\-1}  & \makecell{6\\  - \\ - }  & \makecell{7\\2\\+1}  & \makecell{7\\ - \\ - }  & \makecell{7\\ 1\\ - }  & \makecell{6\\  - \\-1}  & \makecell{7\\  - \\+1}  & \makecell{7\\  - \\ - }  & \makecell{6\\  - \\-1}  & \makecell{6\\  - \\ - }  \\
 \hline
 \hline
 \rotatebox[origin=c]{90}{C. Mode} & \makecell{Point \\ Line \\ Plane} & \makecell{4\\  - \\ - }  & \makecell{3\\  - \\ - }  & \makecell{3\\  - \\ - }  & \makecell{4\\  - \\-}  & \makecell{4\\  - \\ - }  & \makecell{3\\  - \\-}  & \makecell{3\\  - \\ - }  & \makecell{3\\  - \\ - }  & \makecell{4\\  - \\-}  & \makecell{3\\  1 \\-}  & \makecell{2\\  1 \\-}  & \makecell{2\\  1 \\-}  & \makecell{2\\  1 \\-}  & \makecell{2\\ 2\\-}  & \makecell{1\\ 2\\-}  & \makecell{1\\ 1\\1}  & \makecell{1\\ 1\\1}  & \makecell{1\\ 1\\1}  & \makecell{1\\ 1\\1}  & \makecell{1\\ 1\\1}  & \makecell{1\\ 2\\1}  & \makecell{2\\ 1\\1}  & \makecell{2\\ 1 \\ 1 }  & \makecell{2\\ 1 \\ 1 }  & \makecell{2\\ 1 \\ 1 }  & \makecell{2\\ 1\\ 1 }  & \makecell{1\\  1 \\1}  & \makecell{1\\  - \\2}  & \makecell{1\\  - \\ 2 }  & \makecell{-\\  - \\2}  & \makecell{-\\  - \\ 2 }  \\
 \hline
\end{tabular}
\end{center}
\end{table*}

\subsection{Experiment 2: Crouch to Kneel, Simulated with Hardware Validation}

We deployed a modified version of the keyframe sequence in Experiment 1 to a physical Valkyrie robot. The sequence starts in the same configuration but only has a front block, as forearm contact on the left block was not possible due to exposed wiring. The same three initial events are present as in Experiment 1: place right hand, place left hand and place right foot. As shown in Fig. \ref{fig:results_combo}, triple support between two knees and a single hand contact were achieved while placing a hand on the blocks (keyframe 8). Additionally, line contact on the right foot was made while placing the foot down (keyframe 19). The trajectory is tracked at 4s per keyframe using an impedance controller which modulates gains based on limb loading (see Appendix) and achieved an average joint tracking of $(0.023\pm 0.009)$\si{\radian}. This experiment was performed by first teleoperating the robot in simulation, then deploying the trajectory to Valkyrie. Deploying to hardware serves as empirical verification of the presented feasibility checks as well as simulation accuracy. Using this framework for live hardware teleoperation will be pursued in future work and was not performed due to hardware safety concerns.


\subsection{Experiment 3: Bracing Against a Wall, Simulated}

\begin{figure}
    \centering
    \includegraphics[width=\columnwidth]{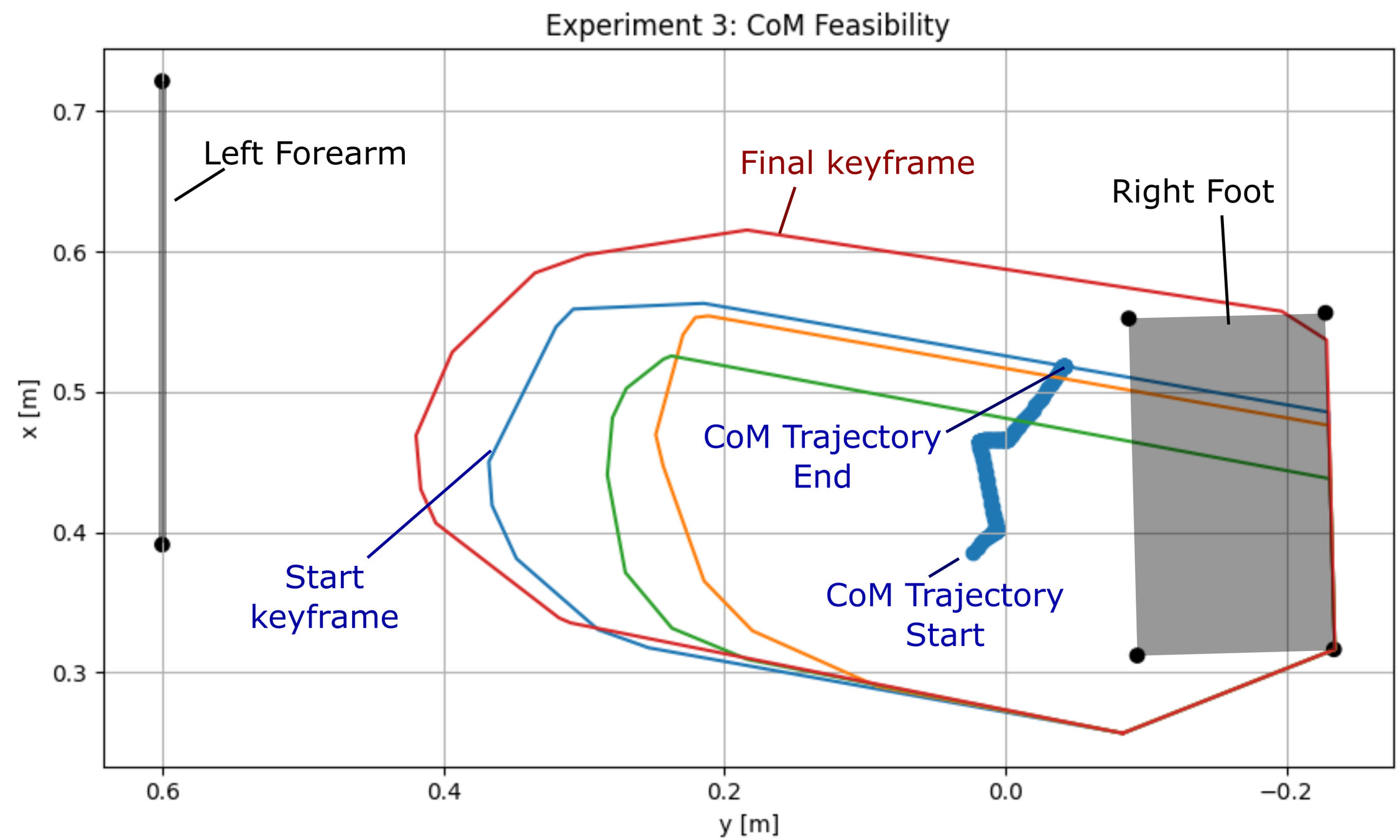}
    \caption{The trajectory for experiment 3 consists of 4 keyframes as the robot is braced against a wall. This top-down view shows the CoM trajectory along with the set of keyframe CoM constraint regions. The regions with blue, orange, green and red outlines correspond to the feasible CoM regions in keyframes 1-4 respectively.}
    \label{fig:exp3_com}
\end{figure}

For experiment 3, the robot stands on a narrow platform with a wall on the side and 34cm tall obstacle blocking the path. The robot braces against the wall with the left forearm while swinging the left foot over the obstacle (Fig. \ref{fig:results_combo}). The use of the left elbow contact reduces load on the elbow joint from 40-55Nm (without elbow contact) to 15-25Nm (with elbow contact). In this scenario, CoM feasibility plays a crucial role as the robot is swinging its foot (Fig. \ref{fig:exp3_com}). During the motion, the feasible CoM region changes on an order of 10cm as the posture, namely the supporting limb Jacobians, are changing. This highlights a case where CoM feasibility is not intuitive and the operator relies on the automated feasibility checks during teleoperation. The trajectory was executed using a whole-body inverse dynamics controller at 2s per keyframe with an average joint tracking of $(0.036\pm 0.008)$\si{\radian}.

\section{DISCUSSION AND FUTURE WORK}

The presented framework demonstrates a teleoperation strategy designed around maximizing the set of user-commandable contact and kinematic constraints. We demonstrated the effectiveness of this strategy by teleoperating in scenarios that are difficult or impossible without the use of knee/elbow contacts, line contacts and arbitrary taskspace objectives. Through real-time feasibility assessment, the operator was able to successfully teleoperate in scenarios where human intuition is likely insufficient. 

Future work will focus on further decreasing operator burden through a number of improvements. One is through explicit modeling of contact modes. In addition to decreasing contact creation time, this allows enumerating contact mode transitions, such as the plane-to-line transition when during toe-off. Keyframe feasibility and interpolations can be improved by leveraging modern model-predictive control \cite{mastalli2020crocoddyl} to perform real-time dynamic feasibility assessment. Time-optimal approaches \cite{pham2014general} will be explored as a strategy for faster motion execution.

\appendix
\textit{Feasible CoM Region}. Following the approach from Orsolino et al. \cite{orsolino2020feasible}, the extreme feasible planar CoM position $\mathbf{c}_{xy} \in \mathbb{R}^2$ along a query direction $\mathbf{a}\in \mathbb{R}^2$ is computed as:

\begin{equation} \label{eq:feasible_region}
\begin{aligned}
    \max_{\mathbf{c}_{xy},\mathbf{f}} \, \mathbf{a}^T \mathbf{c}_{xy} \;\;\; \textrm{s.t.}   \quad &  \mathbf{A}_f \mathbf{f} + \mathbf{A}_c \mathbf{c}_{xy} = \mathbf{b}_f \\
                                            \quad &   \mathbf{C}_f \mathbf{f} \leq \mathbf{d}_f \\ 
\end{aligned}
\end{equation}

Where $\mathbf{A}_f, \mathbf{A}_c,\mathbf{b}_f,\mathbf{C}_f,\mathbf{d}_f$ are the constraints in Eq. \ref{eq:StaticFeasibility}. The equality constraint enforces static equilibrium and the inequality constraint unilateral contact forces, friction constraints, and actuation limits. When solving Eq. \ref{eq:feasible_region} for a given direction $\mathbf{a}_i$, the optimized CoM value $\mathbf{c}_{xy}^*$ lies on the boundary of the feasible CoM region. By querying in a set of directions, an approximation of the feasible CoM region is determined.

\textit{Static Force Distribution} An optimal static force distribution is computed similarly to Eq. \ref{eq:feasible_region} but with $\mathbf{c}_{xy}$ constrained to the keyframe's current position. 
\begin{equation} \label{eq:ForceDistribution}
\begin{aligned}
    \min_{\mathbf{f}} \; \lVert \mathbf{f} \rVert_2^2 \;\; \textrm{s.t.} \quad &  \mathbf{A}_{f} \mathbf{f} = \mathbf{b}_{f} - \mathbf{A}_c\mathbf{c}_{xy}  \\
                        \quad &  \mathbf{C}_{f} \mathbf{f} \leq \mathbf{d}_f
\end{aligned}
\end{equation}

\textit{Valkyrie Impedance Gains} Tab. \ref{tab:impedance_gains} is the set of loaded joint stiffness and damping values used for Experiment 2. A limb is considered loaded when any link contains a contact anchor. Unloaded gains are computed by reducing the loaded gain by a factor of 0.65.

\renewcommand{\arraystretch}{1.0}
\begin{table}[h!]
\centering
\captionof{table}{Hardware Loaded Joint Impedance Gains} \label{tab:impedance_gains} 
\begin{tabular}{ c c c } 
 \hline
  Joint & Stiffness & Damping \\ 
     &  [\si{\newton\meter\per\radian}] & [\si{\newton\meter\second\per\radian}] \\ 
 \hline
Sh. Pitch & 1600 & 32  \\
Sh. Roll & 1750 & 50  \\
Sh. Yaw & 550 & 12  \\
El. Pitch & 900 & 30  \\
Hip Yaw & 1600 &  32   \\
Hip Roll & 2000 & 55  \\
Hip Pitch & 2000 & 55  \\
Knee Pitch & 2000 & 60 \\
Ankle Roll & 800 &  10  \\
Ankle Pitch & 800  & 10  \\
Spine Yaw & 1750 & 65  \\
Spine Pitch & 1500 & 45  \\
Spine Roll & 1250  & 55  \\
 \hline
\end{tabular}
\end{table}

\printbibliography

\balance

\end{document}